%% file: latex/acl_latex.tex
\documentclass[11pt]{article}

\usepackage[preprint]{acl}

\usepackage{times}
\usepackage{latexsym}
\usepackage{booktabs}
\usepackage{multirow}
\usepackage[table]{xcolor}  
\usepackage{graphicx}
\usepackage{listings}
\usepackage{subcaption}
\usepackage{hyperref}
\usepackage{booktabs}
\newcommand{\rev}[1]{\textcolor{red}{#1}}

\renewcommand{\rev}[1]{#1}

\usepackage[T1]{fontenc}
\usepackage[utf8]{inputenc}

\usepackage{microtype}

\usepackage{inconsolata}

\title{Strong Reasoning Isn’t Enough: \\ Evaluating Evidence Elicitation in Interactive Diagnosis}

\author{
\textbf{Zhuohan Long}\textsuperscript{1},
\textbf{Zhijie Bao}\textsuperscript{1,2},
\textbf{Zhongyu Wei}\textsuperscript{1,2}\thanks{Corresponding author.}
\\
\textsuperscript{1}School of Data Science, Fudan University \\
\textsuperscript{2}Shanghai Innovation Institute
\\
\texttt{zhlong24@m.fudan.edu.cn, zjbao24@m.fudan.edu.cn, zywei@fudan.edu.cn}
}

\begin{document}
\maketitle
\begin{abstract}
Interactive medical consultation requires an agent to proactively elicit missing clinical evidence under uncertainty.
Yet existing evaluations largely remain static or outcome-centric, neglecting the evidence-gathering process.
In this work, we propose an interactive evaluation framework that explicitly models the consultation process using a simulated patient and a \rev{simulated reporter} grounded in atomic evidences.
Based on this representation, we introduce Information Coverage Rate (ICR) to quantify how completely an agent uncovers necessary evidence during interaction.
To support systematic study, we build EviMed, an evidence-based benchmark spanning diverse conditions from common complaints to rare diseases, and evaluate 10 models with varying reasoning abilities.
We find that strong diagnostic reasoning does not guarantee effective information collection, and this insufficiency acts as a primary bottleneck limiting performance in interactive settings.
To address this, we propose REFINE, a strategy that leverages diagnostic verification to guide the agent in proactively resolving uncertainties.
Extensive experiments demonstrate that REFINE consistently outperforms baselines across diverse datasets and facilitates effective model collaboration, enabling smaller agents to achieve superior performance under strong reasoning supervision.
Our code can be found at \href{https://github.com/NanshineLoong/EID-Benchmark}{this URL}.
\end{abstract}

% Introduction
\input{latex/introduction}

% Evaluation Framework
\input{latex/evaluation_framework}

% EviMed-1k Benchmark construction
\input{latex/evimed_benchmark}

% REFINE Strategy
\input{latex/refine_strategy}

% Experiments
\section{Experiments}
\input{latex/experiments/setup}

\input{latex/experiments/ex_overall}

\input{latex/experiments/ex_modes}

\input{latex/experiments/ex_icr_sr}
\input{latex/experiments/ex_complementarity}
\input{latex/experiments/ex_cov_sr}
\input{latex/experiments/ex_turns}
\input{latex/experiments/ex_sanity}

% Related Work
\input{latex/related_work}

% Conclusion
\input{latex/conclusion}

% Limitation
\input{latex/limitation}

\bibliography{latex/custom}

\clearpage
\appendix
\section{Interactive Environment and Evaluation Details}
\input{latex/appendix/interactive_environment}

\section{EviMed Construction Details}
\input{latex/appendix/evimed_construction_details}

\section{Interaction Turns}
\input{latex/appendix/avg_turns}

\section{Latency} 
\input{latex/appendix/avg_turn_time}

\section{Strategy Prompts}
\input{latex/appendix/prompts}

\end{document}

%% file: latex/introduction.tex
\begin{figure}[t]
  \centering
  \includegraphics[width=\linewidth]{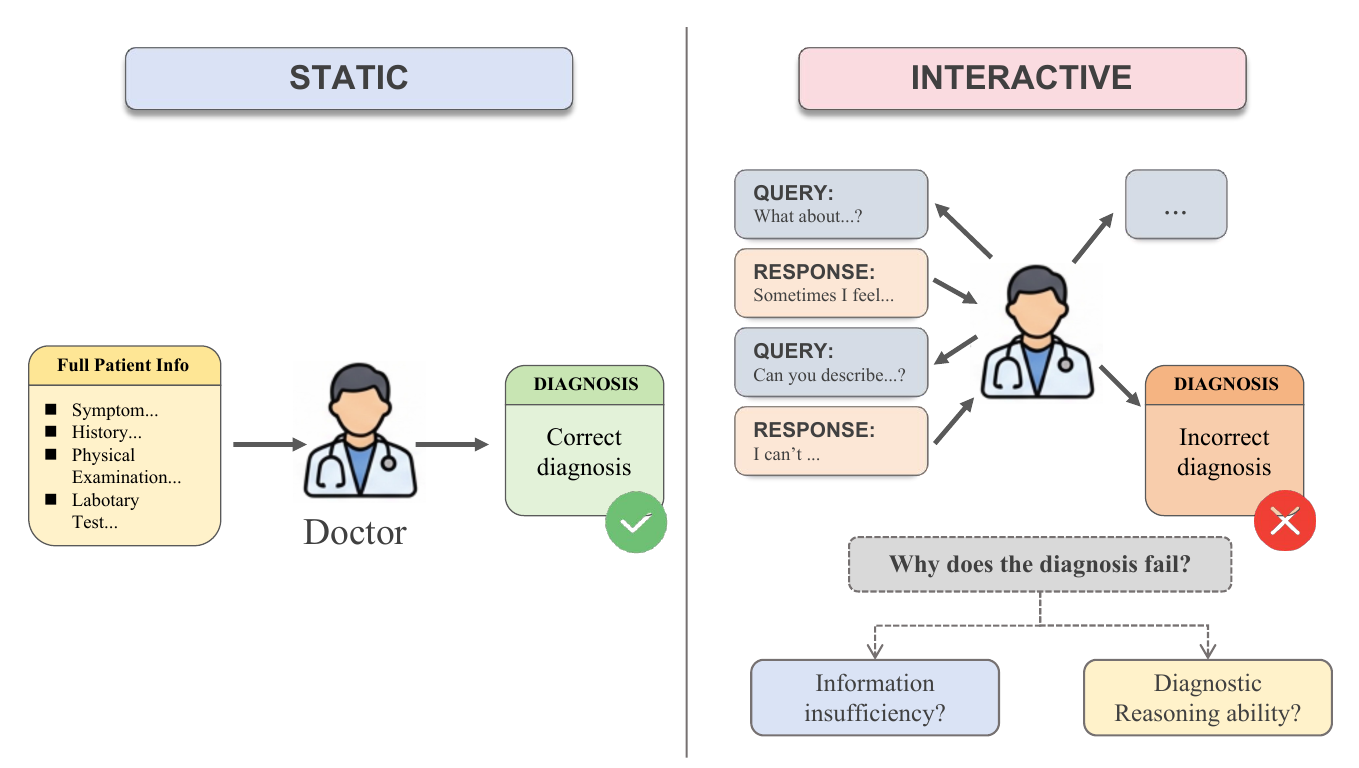}
  \caption{Static evaluation provides full patient information upfront. Interactive diagnosis requires iterative evidence elicitation and may fail due to insufficient information gathering or flawed reasoning.}
  \label{fig:static-interactive}
\end{figure}

\section{Introduction}

% Background: recent advancements of LLM agents
Large Language Models (LLMs) have achieved remarkable progress in recent years, evolving from passive language processors to autonomous agents~\cite{ahn2022can, liu2023agentbench}.
Beyond text generation, these agents demonstrate increasing capabilities in interacting with external environments~\cite{zhou2023webarena, yao2022webshop}, using tools~\cite{schick2023toolformer}, and executing complex workflows~\cite{jimenez2023swe}.
Such advancements suggest that LLM-based agents are becoming proficient at following user instructions to accomplish multi-step tasks.

% Transition: from instruction-following to information-seeking
However, many real-world decision-making scenarios extend beyond simple instruction following.
In these settings, the agent is not provided with all necessary context upfront.
Instead, the agent must actively identify missing information and acquire it through interaction before making a decision.
Consequently, the quality of the final outcome hinges heavily on the agent's ability to gather information effectively under uncertainty.

% Medical consultation as a representative information-seeking task
Medical consultation represents a quintessential instance of such information-seeking scenarios.
In clinical practice, diagnosis is an interactive evidence-gathering process rather than an one-shot prediction task~\cite{meyer2021patient}.
Key evidence, including symptoms, medical history, and examinations, must be actively elicited through patient inquiry or clinical testing.
Therefore, an effective medical agent must proactively ask relevant questions and decide when sufficient evidence has been collected to support a reliable diagnosis.

% Gap: Limitation of existing static and interactive evaluation
Despite this interactive nature (Fig.~\ref{fig:static-interactive}), most existing evaluations~\cite{nori2023can, chen2023meditron, lievin2024can, wu2024pmc, singhal2025toward} focus on static settings where all patient information is provided to the LLM in advance.
While recent studies have begun to explore interactive diagnosis, they still primarily assess the final diagnostic accuracy.
This outcome-oriented evaluation overlooks the evidence-elicitation process, leaving it unclear whether the agent can efficiently and systematically gather the information required for a correct diagnosis.

% Proposal: Framework, Atomic Evidence, and ICR
In this work, we argue that information collection ability should be treated as a first-class evaluation target for medical agents.
To this end, we propose an interactive evaluation framework that explicitly models the consultation process using a simulated patient and a \rev{simulated reporter} by leveraging the generative capabilities of language models~\cite{park2023generative, du2024llms}.
We specifically represent all clinical information within these modules as atomic evidences, which are defined as minimal and self-contained units of facts.
This granular representation enables us to introduce a new metric, the Information Coverage Rate (ICR).
Unlike traditional success rates, ICR explicitly measures the proportion of necessary evidence successfully revealed by the agent, providing a fine-grained assessment of its active inquiry capabilities.

% Benchmark: EviMed
To support this evaluation framework, we construct EviMed, a new benchmark for interactive medical consultation.
We transform existing medical datasets into the evidence-based format through an automated construction process.
The resulting benchmark covers a diverse range of scenarios, spanning from common medical inquiries to challenging rare disease diagnoses that require extensive evidence accumulation.

% Key Findings: Performance drop and insights
We evaluate 10 LLMs of varying diagnostic reasoning ability on EviMed, revealing a performance disparity between static and interactive settings.
On average, diagnostic success rates drop by approximately 20\% when agents are required to actively collect information, with even sharper declines observed in rare disease scenarios.
Moreover, our results show that strong diagnostic reasoning alone does not guarantee effective information acquisition. Insufficient information collection during interaction appears to be the main bottleneck underlying performance degradation.

% Method: REFINE (Feedback-driven strategy) and its results
To address these challenges, we propose REFINE (Reasoning-Enhanced Feedback for INformation Elicitation).
It employs a Diagnosis Verifier to examine whether a generated Diagnosis hypothesis is fully grounded in the collected evidences, guiding the agent to resolve uncertainties.
Extensive experiments demonstrate that this strategy effectively mitigates the performance degradation in interactive settings, yielding substantial improvements in both information coverage and diagnostic success rates across diverse models. Furthermore, we find that REFINE enables effective collaboration between heterogeneous models, allowing smaller, inquisitive agents to achieve superior performance when supervised by strong reasoning models.

% Contributions
Overall, our contributions are threefold:

(1) We propose an interactive evaluation framework for evidence collection and introduce the Information Coverage Rate (ICR) metric, which formalizes information collection as a measurable objective grounded in atomic evidence units.

(2) We construct EviMed, a comprehensive diagnostic benchmark, and conduct a systematic evaluation of 10 models with varying diagnostic capabilities, revealing that strong diagnostic reasoning does not guarantee effective information collection, which emerges as a key performance bottleneck.

(3) We introduce REFINE, a feedback-driven strategy that leverages diagnostic verification to guide the interaction. It effectively mitigates the performance degradation in interactive settings and supports heterogeneous model collaboration, enabling smaller models to excel in inquiry tasks under strong reasoning supervision.

%% file: latex/evaluation_framework.tex
\section{Interactive Evidence Collection Evaluation Framework}

\begin{figure*}[t]
    \centering
    \includegraphics[width=0.80\linewidth]{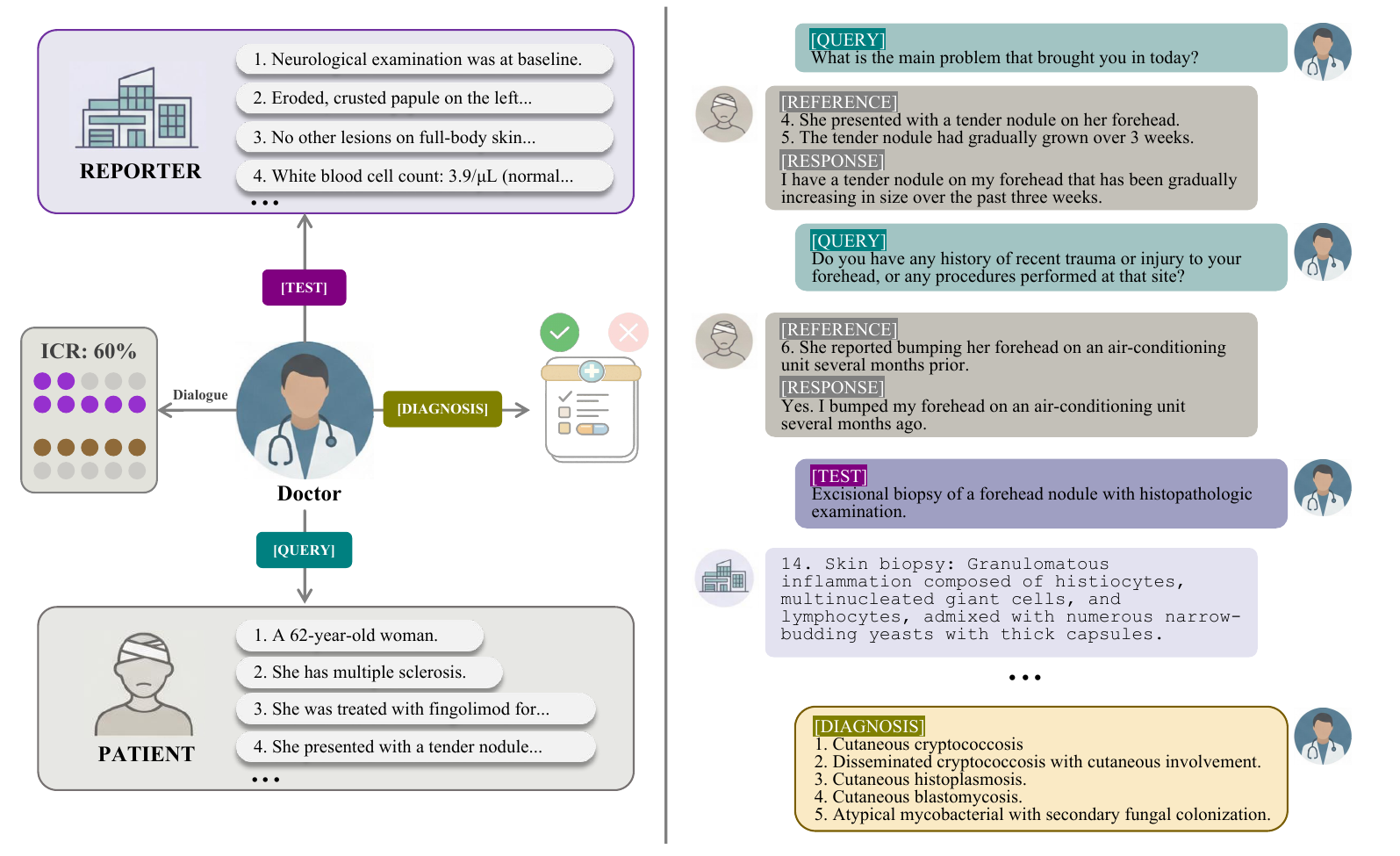}
    \caption{Interactive Evaluation Framework. \textbf{(Left)} The consultation loop. The doctor agent iteratively interacts with a simulated patient that provides subjective/history-related information and a \rev{simulated reporter} that returns objective examination or laboratory findings. The interaction reveals atomic evidences used for evaluation, where we track \emph{Information Coverage Rate (ICR)} to measure how many relevant evidences have been collected during the dialogue, and assess diagnostic \emph{Success Rate} based on the final diagnosis. \textbf{(Right)} An example consultation trajectory showing queries, evidence-grounded responses, test requests/results, and the final diagnosis.}

    \label{fig:eval_framework}
\end{figure*}

% This paragraph motivates interactive evidence collection and introduces the evaluation setting.
In real clinical settings, relevant information is often incomplete and distributed across multiple sources.
A clinician must determine what information is missing, how to obtain it, and when the collected evidence is sufficient to support a diagnosis.
To model this interactive evidence collection process, we construct a medical consultation environment concentrating on evidence collection shown in Figure~\ref{fig:eval_framework}.

% This paragraph summarizes the consultation loop and clarifies what each role does in interaction.
The environment is composed of three distinct roles, including a simulated patient, a \rev{simulated reporter}, and the doctor agent being evaluated.
The consultation proceeds in multiple turns.
At each turn, the doctor agent chooses an action, including asking the patient a question, requesting a clinical examination, or issuing a diagnosis.
In response, the simulated patient and the \rev{simulated reporter} return information grounded in the case record.

\subsection{Roles}

% This paragraph defines the shared evidence representation used by the roles and the evaluation.
We define the three roles and their modeling assumptions below.

% This paragraph describes the simulated patient and its conservative disclosure policy.
\textbf{Simulated Patient.}
The simulated patient maintains the patient’s personal information in the form of atomic evidences, covering symptoms, history, and related clinical details.
It follows an evidence selection mechanism similar to the Fact-Select approach in~\citet{li2024mediq}, the patient selects the most relevant evidences for a given query and generates a natural language response grounded in them.
Each response is supported by at most two evidences.
If the query is unrelated to any evidence, the patient explicitly indicates uncertainty.

% This paragraph describes the simulated reporter and how it returns objective results.
\textbf{\rev{Simulated Reporter}.}
The \rev{simulated reporter} maintains clinical examination results and laboratory test findings.
For each request, it returns one or more relevant evidences as objective observations.
These results are provided directly without natural language generation.
This component evaluates whether the agent can select appropriate examinations and utilize objective clinical evidence.

% This paragraph defines the doctor agent and frames consultation as a sequential decision making problem.
\textbf{Doctor Agent.}
The doctor agent is the model under evaluation.
During the consultation, it decides its next action among asking a question, requesting a test, or issuing a diagnosis.
This requires the agent to determine whom to interact with, how to formulate queries, and when to terminate the interaction.
Therefore, the consultation process is modeled as a sequential decision-making problem under incomplete information.

\subsection{Information Coverage Rate}

% This paragraph introduces ICR as a metric for evidence collection completeness.
To evaluate the agent’s ability to collect relevant information, we propose Information Coverage Rate (ICR).
ICR measures the proportion of evidences that are successfully collected by the agent through interaction.
It reflects how thoroughly the agent explores the evidence space required for diagnosis.

% This paragraph provides the formal definition of ICR.
Formally, let $E$ denote the set of all relevant evidences for a given case.
Let $\hat{E}$ denote the set of evidences collected by the agent during the consultation.
ICR is defined as
\[
\mathrm{ICR} = \frac{|\hat{E} \cap E|}{|E|}.
\]

% This paragraph explains why ICR is directly computable in the proposed framework and how it complements accuracy.
As both patient responses and test results are grounded in atomic evidences, ICR is directly computable from the evidence revealed during interaction.
Together with diagnostic success rate, ICR provides a complementary view by separating evidence collection completeness from final diagnostic correctness.

%% file: latex/evimed_benchmark.tex
\section{EviMed Benchmark Construction}

% Motivate benchmark construction by linking framework requirements to dataset limitations.
Most existing medical datasets present complete case narratives without explicit atomic evidence structure.
This makes it difficult to support selective evidence disclosure and compute information collection coverage.
To address this gap, we construct EviMed, an evidence-based benchmark for interactive medical consultation evaluation.

\subsection{Source Datasets}

% Overview of dataset composition and unified sampling strategy.
EviMed integrates five complementary data sources covering general medicine, specialty diagnosis, complex multi-specialty reasoning, rare diseases, and real-world clinical records.
We sample two hundred cases from each source, yielding the EviMed-1K benchmark, which spans a wide range of diagnostic settings. 
The five data sources are described as follows:

\textbf{AgentClinic-MedQA~\cite{schmidgall2024agentclinic}} is adapted from USMLE-style medical examination cases and rewritten into consultation-oriented scenarios.
It covers a wide range of common diseases and clinical conditions.
We use this source to evaluate general diagnostic reasoning.

\textbf{Derm~\cite{johri2024craft}} focuses on dermatological diagnosis and emphasizes fine-grained descriptions.
It contains both publicly available cases and clinician-authored cases with similar structures.
We include the full set to evaluate detailed symptom inquiry in a specialized domain.

\textbf{DiagnosisArena~\cite{zhu2025diagnosisarena}} is constructed from real-world case reports published in top-tier medical journals.
The cases require complex diagnostic reasoning across multiple clinical specialties.
We use it to assess information collection in challenging diagnostic scenarios.

\textbf{RareArena~\cite{zhao2025rarearena}} is built from publicly available case reports in PubMed Central (PMC) and covers a wide range of rare diseases, which involve limited prior knowledge and ambiguous symptom presentations.
We sample cases according to disease frequency to encourage diversity across different levels of rarity.

\textbf{ClinicalBench~\cite{yan2024clinicallab}} is derived from real electronic medical records containing both structured and unstructured information.
It covers cases from multiple clinical departments and a broad set of disease categories.
We sample cases to ensure coverage across disease types.

\subsection{Automatic Construction}

% Describe the evidence construction pipeline and its alignment with the framework.
For each selected case, we transform the original case description into an evidence-based representation.
We separate patient basic information from examination-related information, and then decompose each part into non-overlapping atomic evidences, where each evidence corresponds to a minimal and self-contained clinical fact.
This conversion is performed automatically using GPT-5-mini.

% Paragraph purpose: describe the constructed outputs and benchmark statistics.
After construction, each case is associated with a set of patient evidences and a set of examination evidences.
These evidences serve as the information sources accessed during interaction by the simulated patient and the \rev{simulated reporter}.
Table~\ref{tab:evimed_stats} summarizes the benchmark statistics, including the average number of patient evidences and examination evidences per case in each source dataset.
The number of atomic evidences varies across sources, reflecting differences in case complexity and information density.

% Paragraph purpose: validate information preservation after construction.
In Section~\ref{ex:sanity_check}, we verify that the automatic construction process preserves diagnostic information.

\begin{table}[t]
    \centering
    \small
    \setlength{\tabcolsep}{6pt} % 缩小列间距（默认 6pt，可再减）
    \caption{Statistics of EviMed-1K.}
    \label{tab:evimed_stats}
    \resizebox{\columnwidth}{!}{%
    \begin{tabular}{lccc}
    \hline
    Dataset & Data Size & Avg Pat. Evi. & Avg Exam Evi. \\
    \hline
    AgentClinic-MedQA & 200 & 14.87 & 12.73 \\
    Derm               & 200 & 7.31  & 2.87  \\
    DiagnosisArena     & 200 & 8.04  & 12.82 \\
    RareArena          & 200 & 17.11 & 17.72 \\
    ClinicalBench      & 200 & 21.76 & 21.14 \\
    \hline
    \end{tabular}}
\end{table}

%% file: latex/refine_strategy.tex
\section{REFINE: Feedback-Driven Evidence Collection}
% We frame interactive consultation as evidence acquisition under diagnostic uncertainty and state the two challenges.
In interactive medical consultation, the agent must make diagnostic decisions under incomplete and evolving evidence.
This setting introduces two tightly coupled challenges.
First, the agent must determine which information to elicit next in order to efficiently reduce the diagnostic uncertainty.
Second, it must decide when the accumulated evidence is sufficient to support a reliable diagnosis, rather than terminating the consultation prematurely.

% We introduce REFINE and summarize the key idea that diagnosis attempts are used to produce feedback that guides what to ask and when to stop.

% We provide a high-level system view and clarify the control flow around stopping, evidence organization, and the diagnose and verify stages.

\begin{figure*}[t]
    \centering
    \includegraphics[width=0.95\linewidth]{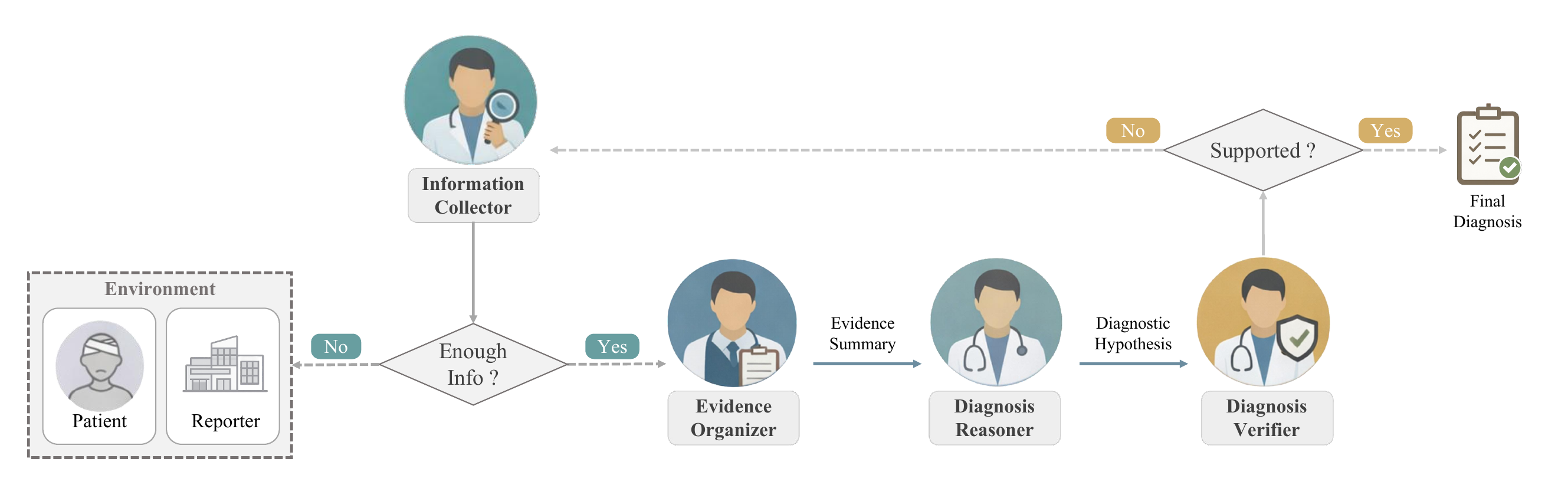}
    \caption{Overview of the REFINE Strategy.}
    \label{fig:refine}
\end{figure*}

To address these challenges, we propose Reasoning-Enhanced Feedback for INformation Elicitation (REFINE), a feedback-driven strategy for evidence collection. As illustrated in Figure~\ref{fig:refine}, REFINE consists of an Information Collector, an Evidence Organizer, a Diagnosis Reasoner and a Diagnosis Verifier.
The Information Collector interacts with the consultation environment across multiple turns. At each turn, it assesses whether the currently collected information is sufficient, decides whether to continue acquiring evidence, or terminates the interaction to make a diagnosis.
When the collector stops, the Evidence Organizer consolidates the collected findings into a structured evidence summary.

% We define the diagnose and verify steps and clarify the role of internal hypotheses versus final diagnostic output.
Given the organized evidence summary, the Diagnosis Reasoner produces a diagnostic hypothesis.
The Diagnosis Verifier then checks whether the hypothesis is fully supported by the available evidence.
If the verifier detects the evidence is insufficient, it provides explicit feedback identifying missing information and unresolved uncertainties, which is sent back to the Information Collector to resume the interaction phase and guide subsequent evidence acquisition steps.

% We describe the iterative feedback loop and termination criteria and explicitly connect them to the two challenges.

This loop continues until the verifier finds that the hypothesis is sufficiently supported by collected evidence or the interaction reaches a maximum turn limit.
This design separates an internal hypothesis used for probing the evidence state from the final diagnostic output.
As a result, the feedback specifies what to collect next, and the absence of critical evidence gaps provides a natural criterion for when to stop.

%% file: latex/experiments/setup.tex
\subsection{Setup}

% Experimental setting and evaluation protocol
We evaluate different models and methods under the interactive evidence collection framework with a maximum of 16 interaction turns.
We consider the following methods for comparison:

% Upper bound reference
\textbf{Upper Bound} uses a static full-information setting where all patient information and examination results are provided upfront.
We prompt the model to generate intermediate reasoning before producing the final diagnosis, establishing a performance upper bound where active information acquisition is not required.

% Basic interactive baseline
\textbf{Baseline} represents a standard interactive setting where a single doctor agent interacts directly with the environment.
At each turn, the model determines whether to ask a question, request a specific examination, or terminate the session to output a final diagnosis.

% Reasoning-augmented interaction
\textbf{ReAct~\cite{yao2022react}} augments the baseline by enforcing an explicit Thought-Act cycle during the consultation.
Before taking any external action, the agent must generate a reasoning trace to analyze the current clinical state and justify its next move, thereby improving decision-making.

% Summarize-then-diagnose baseline
\textbf{Summarized-Conversation (SC)~\cite{johri2024craft}} decouples information gathering from diagnosis.
It first conducts a full multi-turn consultation to collect evidence, then summarizes the entire interaction history into a structured format.
The final diagnosis is produced solely based on this consolidated summary rather than turn-level context.

% Evidence-based feedback strategy
\textbf{REFINE} is our proposed feedback-driven strategy designed to optimize evidence collection.
It utilizes a diagnostic verification mechanism to assess the sufficiency of collected information, providing the agent with explicit feedback to guide subsequent inquiry steps and proactively resolve remaining uncertainties.

% Model backbones and selection
We evaluate a diverse set of language models spanning different scales and domain specializations.
The evaluated models include GPT-5~\cite{openai2025chatgpt5}, GPT-5-mini, DeepSeek-v3.2~\cite{liu2025deepseek}, GLM-4.6~\cite{zai2025glm46}, Qwen2.5-72B~\cite{hui2024qwen2}, Qwen2.5-32B, Qwen2.5-7B, Qwen2.5-3B, Llama-3.1-8B-Instruct~\cite{dubey2024llama}, and Meditron3-8B~\cite{sallinen2025llama}.

%% file: latex/experiments/ex_overall.tex
\subsection{Static vs. Interactive Evaluation}
\label{sec:ex_static_interactive}

% Motivation and evaluation setup
We compare static and interactive evaluation to assess whether strong full-information reasoning performance transfers to realistic consultations that require evidence acquisition.
Specifically, we report the Success Rate under the static full-information upper bound setting and report both ICR and SR under the interactive baseline.
Results are summarized in Table~\ref{tab:main_results}.

% Overall degradation from static to interactive evaluation
\rev{Across datasets and models, SR decreases by approximately 20\% on average when moving from static to interactive evaluation. This degradation is more pronounced on the more challenging DiagnosisArena and RareArena datasets. Even for the strongest model, GPT-5, performance drops substantially in the interactive setting, indicating that strong static reasoning does not directly translate to effective interactive decision-making.}

% Model differences and capability emphasis
\rev{Interestingly, some models exhibit larger performance drops than others. For example, GPT-5-mini originally achieves a stronger static upper bound than DeepSeek-v3.2 and GLM-4.6, but its interactive SR lags behind them. A similar phenomenon is observed for Meditron3-8B. Although it is fine-tuned on clinical data based on Llama-3.1-8B, it shows a larger performance degradation than its base model.}

\rev{We further observe that models with larger degradation, such as GPT-5-mini and Meditron3-8B, also exhibit relatively low ICR. This suggests that insufficient or inefficient information acquisition may be a key factor limiting their diagnostic reasoning performance in interactive settings.}

\begin{table*}[t]
\caption{Static Upper Bound vs.\ interactive Baseline evaluation. UB denotes SR under static evaluation. ICR and SR are metrics under interactive evaluation framework. For SR, the subscript indicates the percentage decrease relative to UB. \textbf{Bold} values denote the best performance under each metric.}
\label{tab:main_results}
\centering
\setlength{\tabcolsep}{3pt}
\renewcommand{\arraystretch}{1.15}
\resizebox{\textwidth}{!}{%
\begin{tabular}{l*{5}{ccc}}
\toprule
Model & \multicolumn{3}{c}{ClinicalBench} & \multicolumn{3}{c}{Derm} & \multicolumn{3}{c}{DiagnosisArena} & \multicolumn{3}{c}{MedQA} & \multicolumn{3}{c}{RareArena} \\
\cmidrule(lr){2-4} \cmidrule(lr){5-7} \cmidrule(lr){8-10} \cmidrule(lr){11-13} \cmidrule(lr){14-16}
& \cellcolor{gray!15}{UB (\%)$\uparrow$} & ICR (\%)$\uparrow$ & SR (\%)$\uparrow$
& \cellcolor{gray!15}{UB (\%)$\uparrow$} & ICR (\%)$\uparrow$ & SR (\%)$\uparrow$
& \cellcolor{gray!15}{UB (\%)$\uparrow$} & ICR (\%)$\uparrow$ & SR (\%)$\uparrow$
& \cellcolor{gray!15}{UB (\%)$\uparrow$} & ICR (\%)$\uparrow$ & SR (\%)$\uparrow$
& \cellcolor{gray!15}{UB (\%)$\uparrow$} & ICR (\%)$\uparrow$ & SR (\%)$\uparrow$ \\
\midrule
GPT-5
& \cellcolor{gray!15}{64.0} & 35.2 & \ensuremath{47.0_{\scriptsize(-27\%)}}
& \cellcolor{gray!15}{\textbf{93.5}} & 54.9 & \ensuremath{\mathbf{75.0}_{\scriptsize(-20\%)}}
& \cellcolor{gray!15}{\textbf{75.0}} & 55.4 & \ensuremath{\mathbf{47.0}_{\scriptsize(-37\%)}}
& \cellcolor{gray!15}{\textbf{97.5}} & 31.3 & \ensuremath{\mathbf{78.0}_{\scriptsize(-20\%)}}
& \cellcolor{gray!15}{\textbf{75.0}} & 42.7 & \ensuremath{\mathbf{37.0}_{\scriptsize(-51\%)}} \\
GPT-5-mini
& \cellcolor{gray!15}{\textbf{67.5}} & 31.3 & \ensuremath{46.5_{\scriptsize(-31\%)}}
& \cellcolor{gray!15}{84.0} & 51.7 & \ensuremath{57.5_{\scriptsize(-32\%)}}
& \cellcolor{gray!15}{63.0} & 47.2 & \ensuremath{23.0_{\scriptsize(-64\%)}}
& \cellcolor{gray!15}{92.0} & 33.0 & \ensuremath{61.0_{\scriptsize(-34\%)}}
& \cellcolor{gray!15}{68.0} & 35.7 & \ensuremath{15.5_{\scriptsize(-77\%)}} \\
DeepSeek-v3.2
& \cellcolor{gray!15}{63.0} & \textbf{46.7} & \ensuremath{\mathbf{51.0}_{\scriptsize(-19\%)}}
& \cellcolor{gray!15}{80.5} & \textbf{77.0} & \ensuremath{65.0_{\scriptsize(-19\%)}}
& \cellcolor{gray!15}{56.5} & \textbf{63.4} & \ensuremath{21.5_{\scriptsize(-62\%)}}
& \cellcolor{gray!15}{92.5} & 44.3 & \ensuremath{70.5_{\scriptsize(-24\%)}}
& \cellcolor{gray!15}{61.5} & \textbf{49.1} & \ensuremath{24.5_{\scriptsize(-60\%)}} \\
GLM-4.6
& \cellcolor{gray!15}{60.5} & 32.6 & \ensuremath{45.0_{\scriptsize(-26\%)}}
& \cellcolor{gray!15}{78.5} & 65.7 & \ensuremath{66.0_{\scriptsize(-16\%)}}
& \cellcolor{gray!15}{52.0} & 52.1 & \ensuremath{23.0_{\scriptsize(-56\%)}}
& \cellcolor{gray!15}{86.0} & 36.6 & \ensuremath{72.0_{\scriptsize(-16\%)}}
& \cellcolor{gray!15}{59.0} & 37.8 & \ensuremath{21.5_{\scriptsize(-64\%)}} \\
Qwen2.5-72B
& \cellcolor{gray!15}{64.5} & 40.8 & \ensuremath{43.5_{\scriptsize(-33\%)}}
& \cellcolor{gray!15}{59.0} & 67.5 & \ensuremath{43.0_{\scriptsize(-27\%)}}
& \cellcolor{gray!15}{24.5} & 55.5 & \ensuremath{10.0_{\scriptsize(-59\%)}}
& \cellcolor{gray!15}{75.5} & 41.7 & \ensuremath{54.0_{\scriptsize(-29\%)}}
& \cellcolor{gray!15}{32.0} & 41.6 & \ensuremath{10.5_{\scriptsize(-67\%)}} \\
Qwen2.5-32B
& \cellcolor{gray!15}{60.0} & 39.0 & \ensuremath{48.0_{\scriptsize(-20\%)}}
& \cellcolor{gray!15}{49.5} & 71.6 & \ensuremath{35.0_{\scriptsize(-29\%)}}
& \cellcolor{gray!15}{20.0} & 57.0 & \ensuremath{8.0_{\scriptsize(-60\%)}}
& \cellcolor{gray!15}{76.5} & 37.3 & \ensuremath{52.0_{\scriptsize(-32\%)}}
& \cellcolor{gray!15}{29.0} & 43.0 & \ensuremath{6.0_{\scriptsize(-79\%)}} \\
Qwen2.5-7B
& \cellcolor{gray!15}{52.0} & 41.0 & \ensuremath{35.5_{\scriptsize(-32\%)}}
& \cellcolor{gray!15}{37.5} & 71.9 & \ensuremath{20.0_{\scriptsize(-47\%)}}
& \cellcolor{gray!15}{13.0} & 53.1 & \ensuremath{8.0_{\scriptsize(-39\%)}}
& \cellcolor{gray!15}{60.5} & 43.5 & \ensuremath{43.0_{\scriptsize(-29\%)}}
& \cellcolor{gray!15}{19.5} & 41.2 & \ensuremath{5.0_{\scriptsize(-74\%)}} \\
Llama-3.1-8B
& \cellcolor{gray!15}{37.0} & 40.5 & \ensuremath{34.5_{\scriptsize(-7\%)}}
& \cellcolor{gray!15}{38.5} & 74.4 & \ensuremath{25.5_{\scriptsize(-34\%)}}
& \cellcolor{gray!15}{14.0} & 57.3 & \ensuremath{6.0_{\scriptsize(-57\%)}}
& \cellcolor{gray!15}{67.0} & \textbf{47.8} & \ensuremath{57.5_{\scriptsize(-14\%)}}
& \cellcolor{gray!15}{23.0} & 45.0 & \ensuremath{9.0_{\scriptsize(-61\%)}} \\
Meditron3-8B
& \cellcolor{gray!15}{42.5} & 23.4 & \ensuremath{20.0_{\scriptsize(-53\%)}}
& \cellcolor{gray!15}{43.0} & 39.7 & \ensuremath{19.0_{\scriptsize(-56\%)}}
& \cellcolor{gray!15}{10.5} & 33.7 & \ensuremath{2.5_{\scriptsize(-76\%)}}
& \cellcolor{gray!15}{68.0} & 28.7 & \ensuremath{20.5_{\scriptsize(-70\%)}}
& \cellcolor{gray!15}{25.5} & 23.2 & \ensuremath{2.5_{\scriptsize(-90\%)}} \\
Qwen2.5-3B
& \cellcolor{gray!15}{38.0} & 33.8 & \ensuremath{30.0_{\scriptsize(-21\%)}}
& \cellcolor{gray!15}{24.0} & 74.1 & \ensuremath{11.5_{\scriptsize(-52\%)}}
& \cellcolor{gray!15}{9.5} & 47.7 & \ensuremath{5.5_{\scriptsize(-42\%)}}
& \cellcolor{gray!15}{49.5} & 43.7 & \ensuremath{35.5_{\scriptsize(-28\%)}}
& \cellcolor{gray!15}{9.5} & 35.1 & \ensuremath{2.0_{\scriptsize(-79\%)}} \\
\bottomrule
\end{tabular}%
}
\end{table*}

%% file: latex/experiments/ex_modes.tex
\begin{table*}[t]
\centering
\small
\setlength{\tabcolsep}{3.5pt}
\renewcommand{\arraystretch}{1.1}
\caption{Comparison of representative models under different interactive strategies across datasets. \textbf{Bold} indicates the best performance for the same model on the same dataset.}
\label{tab:method_comparison}
\resizebox{\textwidth}{!}{%
\begin{tabular}{ll*{4}{cc}}
\toprule
\multirow{2}{*}{Dataset} & \multirow{2}{*}{Model} &
\multicolumn{2}{c}{Baseline} & \multicolumn{2}{c}{ReAct} & \multicolumn{2}{c}{SC} & \multicolumn{2}{c}{REFINE} \\
\cmidrule(lr){3-4}\cmidrule(lr){5-6}\cmidrule(lr){7-8}\cmidrule(lr){9-10}
& & ICR (\%)$\uparrow$ & SR (\%)$\uparrow$ & ICR (\%)$\uparrow$ & SR (\%)$\uparrow$ & ICR (\%)$\uparrow$ & SR (\%)$\uparrow$ & ICR (\%)$\uparrow$ & SR (\%)$\uparrow$ \\
\midrule
\multirow{3}{*}{ClinicalBench} & GPT-5-mini & 31.3 & 46.5 & 33.5 & \textbf{49.5} & 41.5 & 49.0 & \textbf{43.5} & \textbf{49.5} \\
 & Qwen2.5-72B & 40.8 & 43.5 & 38.7 & 42.0 & 43.2 & 49.5 & \textbf{51.1} & \textbf{53.0} \\
 & Qwen2.5-7B & \textbf{41.0} & 35.5 & 29.8 & 33.0 & 32.5 & 35.0 & 35.3 & \textbf{38.5} \\
\midrule
\multirow{3}{*}{Derm} & GPT-5-mini & 51.7 & 57.5 & 59.5 & 62.0 & 70.9 & \textbf{66.5} & \textbf{76.7} & 66.0 \\
 & Qwen2.5-72B & 67.5 & 43.0 & 71.2 & 45.0 & 77.9 & \textbf{45.5} & \textbf{80.8} & 44.0 \\
 & Qwen2.5-7B & \textbf{71.9} & 20.0 & 63.5 & 23.0 & 62.8 & 25.0 & 67.1 & \textbf{28.5} \\
\midrule
\multirow{3}{*}{DiagnosisArena} & GPT-5-mini & 47.2 & 23.0 & 53.9 & 29.5 & 60.8 & 39.5 & \textbf{64.6} & \textbf{42.0} \\
 & Qwen2.5-72B & 55.5 & 10.0 & 58.9 & 12.5 & 63.4 & 15.5 & \textbf{73.8} & \textbf{18.5} \\
 & Qwen2.5-7B & 53.1 & 8.0 & 45.1 & 3.5 & \textbf{53.6} & 10.5 & 50.8 & \textbf{13.0} \\
\midrule
\multirow{3}{*}{AgentClinic-MedQA} & GPT-5-mini & 33.0 & 61.0 & 33.2 & 67.0 & 40.2 & \textbf{69.5} & \textbf{45.5} & 68.0 \\
 & Qwen2.5-72B & 41.7 & 54.0 & 39.5 & 59.0 & 43.6 & 62.0 & \textbf{53.9} & \textbf{64.5} \\
 & Qwen2.5-7B & \textbf{43.5} & 43.0 & 35.8 & 45.0 & 36.0 & 44.5 & 38.6 & \textbf{45.5} \\
\midrule
\multirow{3}{*}{RareArena} & GPT-5-mini & 35.7 & 15.5 & 39.3 & 24.0 & 46.2 & 30.5 & \textbf{51.8} & \textbf{32.0} \\
 & Qwen2.5-72B & 41.6 & 10.5 & 47.4 & 13.0 & 50.1 & 15.5 & \textbf{59.9} & \textbf{17.0} \\
 & Qwen2.5-7B & \textbf{41.2} & 5.0 & 29.8 & 2.5 & 37.0 & 5.5 & 38.2 & \textbf{7.5} \\
\bottomrule
\end{tabular}%
}
\end{table*}

\subsection{Strategy Comparison}
\label{sec:exp_methods}

% Motivation and setup
We compare interaction strategies to assess their effects on ICR and SR.  
We report results for GPT-5-mini, Qwen2.5-72B, and Qwen2.5-7B in Table~\ref{tab:method_comparison}.  

% Effects of ReAct
From the results, ReAct improves both ICR and SR for stronger models such as GPT-5-mini and Qwen2.5-72B.  
However, for the weaker model Qwen2.5-7B, ReAct decreases both ICR and SR.  
This may reflect increased difficulty under longer multi-turn trajectories~\cite{laban2025llms}, which is also suggested by Section~\ref{ex:max_turns}.  

% Effects of SC
In contrast, SC more consistently improves both ICR and SR across the evaluated models. This strategy likely benefits from separating information collection from the final diagnosis, which helps the agent remain focused on evidence acquisition. Moreover, making the diagnosis based on a structured summary may mitigate reasoning degradation in long conversations.

% Effects of REFINE
REFINE achieves the highest ICR across most datasets and models. The improvements are especially pronounced on the challenging DiagnosisArena and RareArena datasets. These results support reasoning-based feedback as an effective mechanism for aligning information collection with downstream diagnostic needs.

%% file: latex/experiments/ex_icr_sr.tex
\begin{figure}[t]
    \centering
    \includegraphics[width=0.95\linewidth]{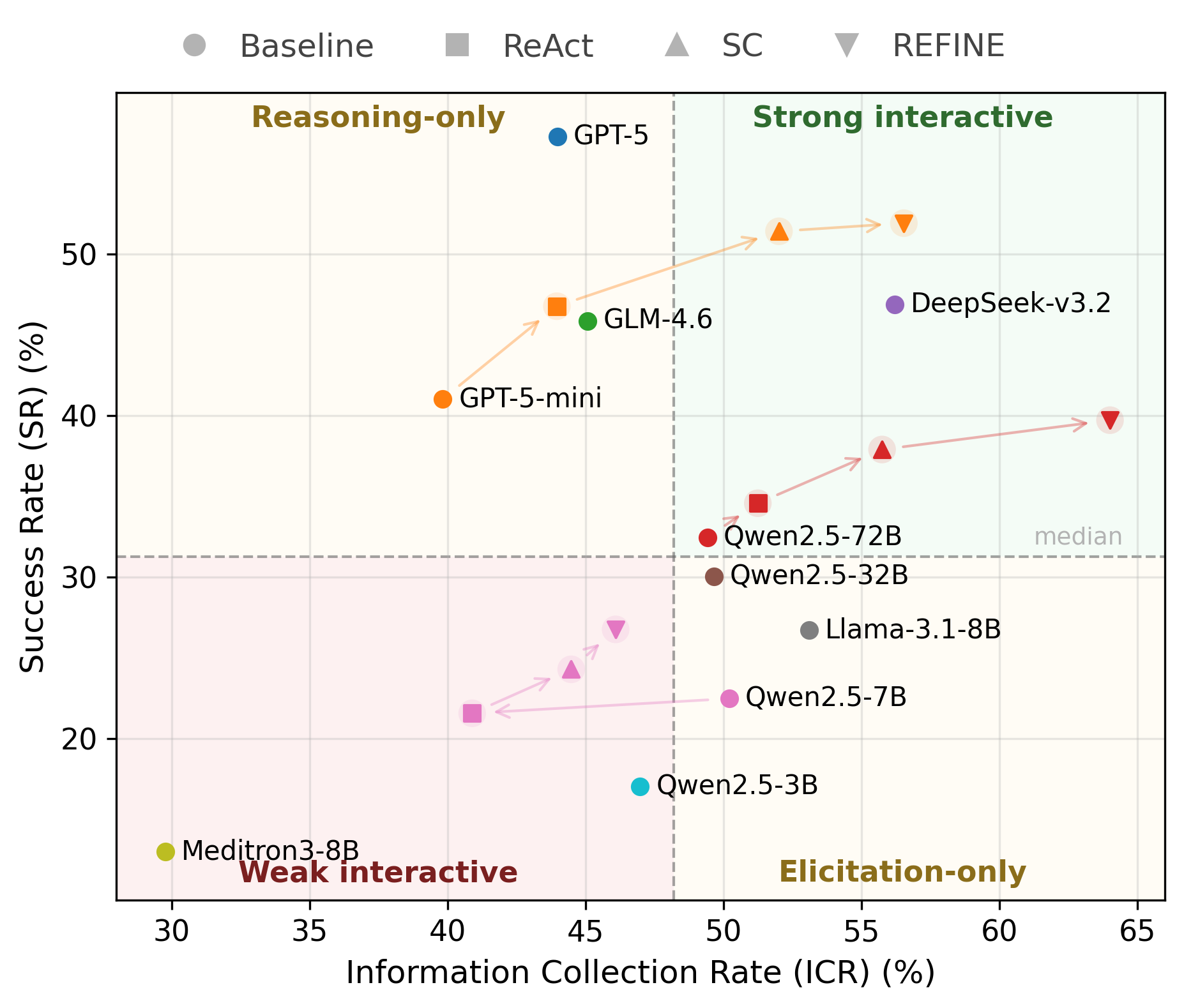}
    \caption{Relationship between average Information Coverage Rate and Success Rate under interactive evaluation, with values averaged over the five datasets.}
    \label{fig:scatter}
\end{figure}

\subsection{Relationship between ICR and SR}
\label{sec:exp_icr_sr}

\rev{To better understand the relationship between evidence acquisition ability and diagnostic performance, we analyze the relationship between Information Collection Rate (ICR) and Success Rate (SR) across models and strategies. We include the interactive results of all models reported in Section~\ref{sec:ex_static_interactive}, as well as the strategy variants for selected representative models in Section~\ref{sec:exp_methods}. The corresponding scatter plots are shown in Figure~\ref{fig:scatter}.}

\rev{We observe that SR generally correlates with ICR across models. For example, performance increases from GPT-5-mini to GLM-4.6 to DeepSeek-v3.2, and from Meditron3-8B to Llama-3.1-8B, in terms of both ICR and SR. However, this relationship is not always consistent. For instance, the GPT-5 series exhibits relatively high SR but comparatively low ICR, whereas the Qwen2.5 series shows high ICR but lower SR. This suggests that diagnostic reasoning ability and evidence elicitation ability are partially decoupled.}

\rev{Specifically, GPT-5 models appear to possess stronger diagnostic reasoning capabilities, enabling them to achieve high performance even with limited information. In contrast, the Qwen2.5 series demonstrates weaker diagnostic reasoning despite effective information collection. Interestingly, within the Qwen2.5 family, model scaling mainly improves SR while yielding marginal gains in ICR, indicating scaling primarily enhances reasoning capacity rather than evidence elicitation ability.}

\rev{From a strategy perspective, we find that most strategies improve SR in accordance with their improvements in ICR, with the exception of ReAct on Qwen2.5-7B, as discussed in Section~\ref{sec:exp_methods}. This further supports a general consistency between ICR and SR, suggesting that enhancing a model’s information acquisition ability is a promising direction for improving overall diagnostic success.}

%% file: latex/experiments/ex_complementarity.tex
\begin{table}[t]
\centering
\caption{Performance comparison of role-aware model pairings under the REFINE strategy.
$M_1 \rightarrow M_2$ denotes using $M_1$ as the Information Collector and $M_2$ as the Organizer, Reasoner and Verifier.}
\label{tab:model_complementarity}
\setlength{\tabcolsep}{3pt}
\renewcommand{\arraystretch}{1.05}
\resizebox{\columnwidth}{!}{%
\begin{tabular}{lrrrrrrrr}
\toprule
\multirow{2}{*}{\textbf{Dataset}}
  & \multicolumn{2}{c}{\textbf{Qwen2.5-7B}}
  & \multicolumn{2}{c}{\textbf{GPT-5-mini}}
  & \multicolumn{2}{c}{\textbf{GPT$\rightarrow$Qwen}}
  & \multicolumn{2}{c}{\textbf{Qwen$\rightarrow$GPT}} \\
\cmidrule(lr){2-3} \cmidrule(lr){4-5} \cmidrule(lr){6-7} \cmidrule(lr){8-9}
  & \textbf{ICR (\%)} & \textbf{SR (\%)}
  & \textbf{ICR (\%)} & \textbf{SR (\%)}
  & \textbf{ICR (\%)} & \textbf{SR (\%)}
  & \textbf{ICR (\%)} & \textbf{SR (\%)} \\
\midrule
ClinicalBench
  & 35.3 & 38.5
  & 43.5 & 49.5
  & 39.9 & 36.5
  & \textbf{52.2} & \textbf{50.5} \\
Derm
  & 67.1 & 28.5
  & 76.7 & 66.0
  & 73.3 & 31.0
  & \textbf{79.5} & \textbf{66.5} \\
DiagnosisArena
  & 50.8 & 13.0
  & 64.6 & 42.0
  & 61.4 & 8.0
  & \textbf{71.7} & \textbf{51.0} \\
MedQA
  & 38.6 & 45.5
  & 45.5 & 68.0
  & 43.8 & 54.0
  & \textbf{54.8} & \textbf{76.5} \\
RareArena
  & 38.2 & 7.5
  & 51.8 & 32.0
  & 46.6 & 6.0
  & \textbf{61.1} & \textbf{46.5} \\
\midrule
\textbf{Average}
  & 46.0 & 26.6
  & 56.4 & 51.5
  & 53.0 & 27.1
  & \textbf{63.9} & \textbf{58.2} \\
\bottomrule
\end{tabular}
}
\end{table}

\subsection{Role-Aware Model Pairing}
\label{sec:model_pairing_refine}

\rev{Motivated by the mismatch between ICR and SR observed for the GPT-5 and Qwen2.5 series in Section~\ref{sec:exp_icr_sr}, we investigate role-aware model pairing within REFINE. Specifically, we assign Qwen2.5-7B as the Information Collector and GPT-5-mini as the Organizer, Reasoner, and Verifier ($Qwen \rightarrow GPT$). For comparison, we also evaluate the reversed role assignment ($GPT \rightarrow Qwen$).}

\rev{As shown in Table~\ref{tab:model_complementarity}, the $Qwen \rightarrow GPT$ configuration achieves the best ICR and SR across all datasets. In particular, it yields substantial improvements in both ICR and SR on DiagnosisArena and RareArena compared to the single GPT-5-mini setting. In contrast, the $GPT \rightarrow Qwen$ configuration consistently underperforms the single GPT-5-mini model in terms of ICR, and even falls below the single Qwen2.5-7B model in SR on three datasets.}

\rev{These results highlight that model mixing is beneficial only when model strengths are aligned with role requirements. They further suggest a cost-effective deployment strategy for REFINE: delegating high-frequency interactions and evidence elicitation to a smaller but inquiry-strong model, while reserving a stronger model for lower-frequency reasoning and verification}

% We further explore the modularity of REFINE by adopting a heterogeneous model setup.
% We utilize Qwen2.5-7B as the Information Collector to interact with the patient, and assign the reasoning-intensive roles (Evidence Organizer, Reasoner, and Verifier) to GPT-5-mini.

% Table~\ref{tab:model_complementarity} summarizes the results.
% The combined strategy consistently outperforms the single-model baselines and achieves the highest average ICR and SR across all datasets.
% Notably, the combined approach yields a 6.7\% average improvement in SR over the homogeneous GPT-5-mini setting.

% We attribute this success to the complementary strengths of the selected models.
% As observed in the Baseline evaluation in Table~\ref{tab:main_results}, Qwen2.5-7B exhibits a strong natural tendency for active inquiry and achieves higher ICR than GPT-5-mini.
% However, it lacks the reasoning precision to effectively convert this evidence into accurate diagnoses.
% Our setup harnesses this inquisitiveness while leveraging the superior reasoning of GPT-5-mini to guide evidence collection and ensure accuracy.

% This synergy is particularly effective on complex benchmarks like DiagnosisArena that require extensive information gathering.
% The results suggest a cost-effective strategy where smaller models handle interaction under the supervision of a strong reasoning engine.

%% file: latex/experiments/ex_cov_sr.tex
\subsection{Information Coverage in Successful vs. Failed Diagnosis}

% Motivation and analysis goal
We examine the association between diagnostic success and information coverage using outcome-conditioned ICR distributions.

\begin{figure}[t]
    \centering
    \includegraphics[width=0.99\linewidth]{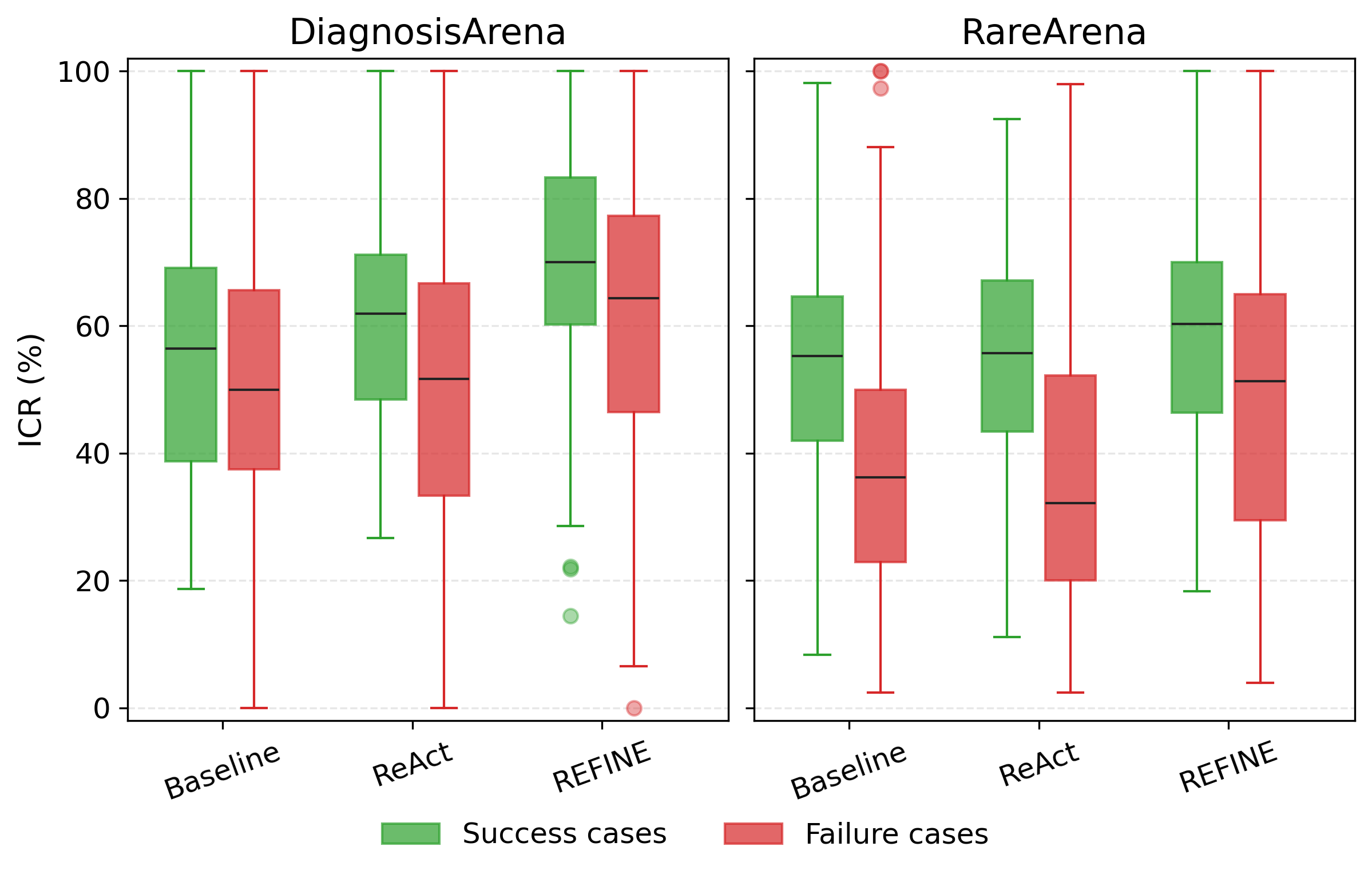}
    \caption{ICR distributions for successful and failed diagnoses on DiagnosisArena and RareArena.}
    \label{fig:outcome_coverage}
\end{figure}

% Experimental setting for success vs failure analysis
We study DiagnosisArena and RareArena, two challenging rare disease benchmarks that typically require extensive information collection.
For each dataset, we aggregate successful and failed cases from the three models used in Section~\ref{sec:exp_methods} and present the distributions of their Information Coverage Rate.
We compare three interaction strategies, Baseline, ReAct, and REFINE.
Figure~\ref{fig:outcome_coverage} summarizes the resulting ICR distributions.

% Main observation
Across both datasets and all strategies, successful cases consistently exhibit higher ICR than failed ones.
These observations indicate that insufficient information coverage is commonly associated with diagnostic failure, supporting ICR as a indicator of diagnostic quality in interactive consultation.

%% file: latex/experiments/ex_turns.tex
\subsection{Effect of Interaction Budget}
\label{ex:max_turns}

\begin{figure}
    \centering
    \includegraphics[width=0.99\linewidth]{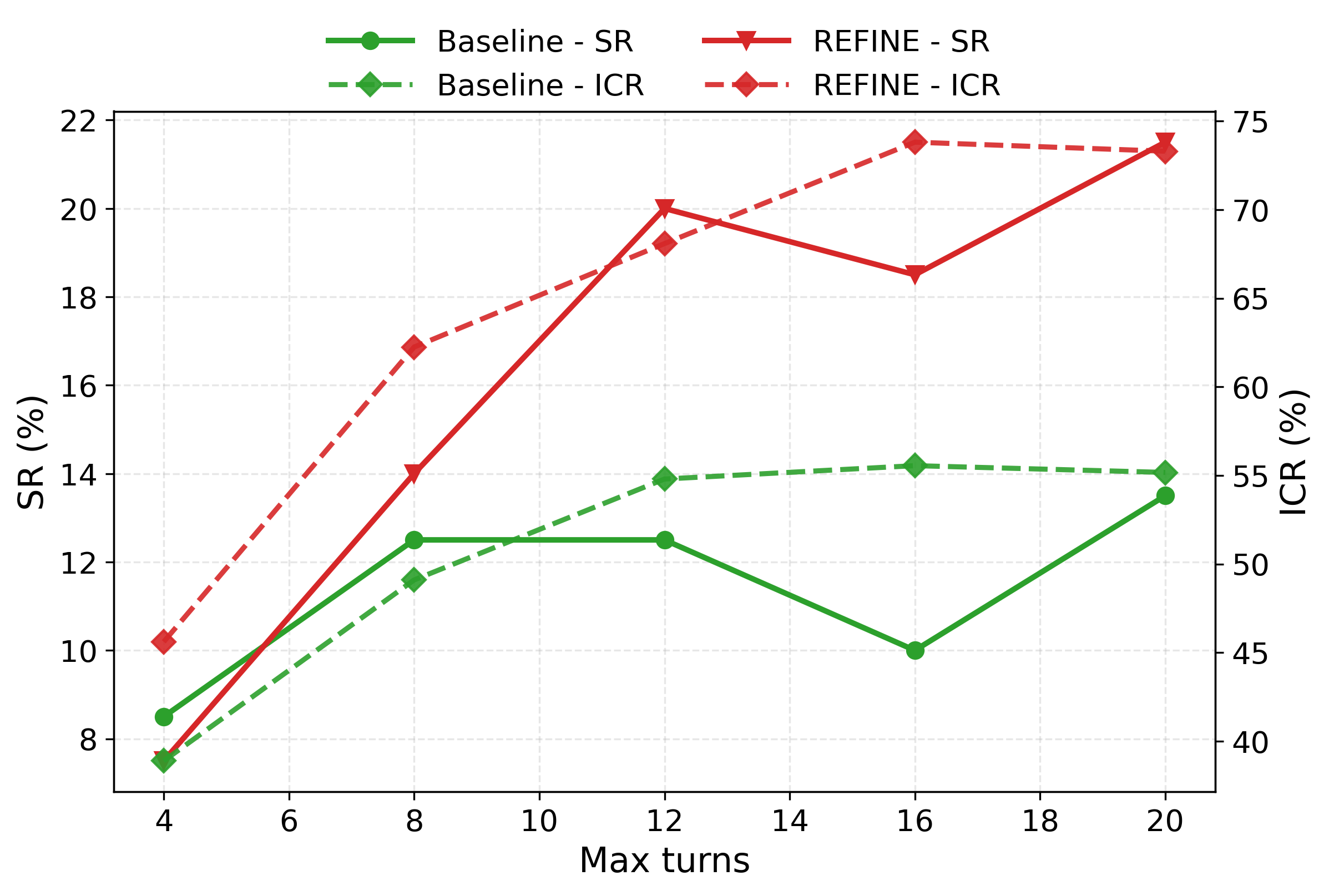}
    \caption{SR and ICR as functions of the maximum number of interaction turns.}
    \label{fig:turn_limit}
\end{figure}

% Experimental setting
We analyze how interaction budget affects ICR and SR.
We vary the maximum number of interaction turns while keeping other settings fixed.
We conduct this study on DiagnosisArena with Qwen2.5-72B, comparing Baseline and REFINE.
Figure~\ref{fig:turn_limit} summarizes the results.

% Main observations with increasing turns
At low budgets, both strategies improve quickly in both ICR and SR.
Both metrics rise sharply in the first few turns.
As the budget grows, the incremental gains taper off.
ICR typically reaches its plateau slightly later than SR.
Comparing the two strategies, 
REFINE sustains higher ICR and SR throughout the range of budgets and saturates later than Baseline.

% Implication
These results indicate that early turns are most effective for acquiring the evidence needed for diagnosis.
After most relevant evidence is collected, additional interaction yields limited benefit and may increase reasoning burden.

% Caption 看需不需要修改

%% file: latex/experiments/ex_sanity.tex
\subsection{Evidence Construction Sanity Check}
\label{ex:sanity_check}

% Motivation of the sanity check
%Our benchmark relies on automatically converting original medical cases into an evidence-based representation.
%This conversion is a necessary step to support interactive evidence collection evaluation.
%However, such conversion may introduce information loss.
We conduct a sanity check to examine whether essential diagnostic information is preserved after evidence construction. 
To ensure essential information is preserved, we compare the diagnostic Success Rate of the original case descriptions against the concatenated constructed evidences in a static evaluation. Table~\ref{tab:evidence_sanity} shows that
the performance differences between original cases and concatenated evidences are small, indicating the information loss introduced by this process is negligible.
%We adopt a static full-information evaluation setting. For each dataset, we compare two input formats. The first uses the original case description. The second concatenates all constructed evidences into a single input. The model performs diagnosis directly based on the provided information in both settings.
%We compare diagnostic SR to assess potential information loss.

% Results and conclusion

\begin{table}[t]
\caption{Diagnostic success rate (\%) under static evaluation using original case descriptions and concatenated evidences.}
\label{tab:evidence_sanity}
\centering
\small
\setlength{\tabcolsep}{8pt}      % 列间距（默认 6pt）
\renewcommand{\arraystretch}{1.1}% 行间距（默认 1.0）
\begin{tabular}{lcc}
\toprule
Dataset & Ori. & Concat. \\
\midrule
DiagnosisArena & 63.0 & 62.5 \\
RareArena      & 68.0 & 69.5 \\
MedQA          & 92.0 & 94.5 \\
ClinicalBench  & 67.5 & 67.5 \\
Derm           & 84.0 & 82.5 \\
\bottomrule
\end{tabular}
\end{table}

%% file: latex/related_work.tex
\section{Related Work}

\paragraph{Task-Oriented Agents}
Early research primarily focused on tool utilization in static scenarios.
In these settings, agents are required to decompose specific user instructions and invoke appropriate APIs or search engines to execute actions~\cite{yao2022react, qin2023toolllm, patil2024gorilla}.

% Multi-turn Interaction in Real-world Environments
Later work~\cite{deng2023mind2web, wang2023mint, yao2022webshop, zhou2023webarena} moved to dynamic environments that require multi-turn interaction, such as web navigation and database manipulation.
Agents must track dialogue state and plan over long horizons to complete tasks reliably.

Another line studies robustness under user-specified policies and evolving constraints in realistic workflows, including retail customer service and flight booking~\cite{yao2024tau, barres2025tau}.
These evaluations prioritize constraint compliance and adaptability during interaction.

% The Critical Pivot: Instruction Following vs. Information Seeking
Most of the above benchmarks assume an instruction-following paradigm where the user states intent and supplies sufficient information.
In many real-world settings, users cannot provide complete information upfront, so agents must form hypotheses and elicit missing evidence through inquiry~\cite{zhu2025ask, mukherjee2024polaris}.
Focusing on the medical consultation setting, our work introduces an interactive evaluation framework that requires agents to proactively gather information throughout the consultation process.

\paragraph{Medical Agent Evaluation}
% Static Evaluation
Medical LLM evaluation has largely relied on static question answering datasets with complete case descriptions, testing knowledge retention and diagnostic reasoning~\cite{jin2021disease, chen2025benchmarking, wang2024cmb, fansi2022ddxplus}.
Multi-agent collaboration can improve reasoning, but it typically remains within full-information inputs and does not require selective evidence discovery~\cite{kim2024mdagents, tang2024medagents, wang2025medagent}.

% Interactive Doctor Patient Simulation
To better reflect clinical practice, recent work simulates doctor patient encounters where agents interact with patients to gather symptoms and request examinations or tests~\cite{schmidgall2024agentclinic, fan2025ai, johri2024craft, almansoori2025medagentsim, bao2025sfmss}.
These environments introduce interaction structure and information asymmetry compared with static benchmarks.

% Process-level Evaluation for Evidence Collection
However, interactive medical evaluations are still commonly scored by final diagnostic accuracy, which weakly captures the quality of the information collection process.
Our work complements this literature by treating evidence collection as a first-class evaluation target and introducing ICR to quantify coverage of necessary atomic evidences during consultation.

%% file: latex/conclusion.tex
\section{Conclusion}

In this work, we revisit medical agent evaluation by shifting the focus from static prediction to interactive evidence collection.
We establish a fine-grained evaluation framework grounded in atomic evidences and construct EviMed to systematically measure the agent's active inquiry capabilities.
Our analysis reveals a critical bottleneck: even models with strong reasoning capabilities often fail to collect sufficient information, leading to a significant performance gap between static and interactive settings.
To address this, we propose REFINE, a strategy that utilizes diagnostic verification to guide the evidence-gathering process.
Experiments demonstrate that REFINE not only improves information coverage and accuracy but also unlocks effective model collaboration, enabling smaller agents to achieve superior results through reasoning supervision.
Ultimately, this work provides a valuable resource for assessing autonomous clinical decision-making and offers a scalable path toward bridging the gap between static knowledge retention and interactive diagnostic reasoning.

%% file: latex/limitation.tex
\section*{Limitations}
\label{sec:limitations}

% The simulated consultation setup may not faithfully reflect real clinical dialogue.
Our evaluation is conducted in a controlled interactive simulator.
This setting may not match the distribution of real patient narratives, clinician behaviors, and institutional constraints.
Simulation can reduce ambiguity and compress the space of plausible follow up trajectories, which can shift the optimal elicitation strategy.
Prior work on simulated patients and multi agent clinical simulators similarly highlights remaining gaps between multi turn interaction and curated or non interactive settings \citep{holderried2024language,fan2025ai,almansoori2025medagentsim}.
Accordingly, our results should be read as comparative performance within this environment rather than a direct measure of clinical readiness.

% Atomic evidence representations may not capture the full complexity of clinical information.
We model clinical information as a set of atomic evidences to enable systematic scoring.
This abstraction omits graded severity, temporal evolution, and dependencies across findings.
Clinical facts also exhibit substantial semantic variability across documentation styles and contexts.
Recent studies on feature and concept extraction from clinical notes suggest that fine grained clinical signal recovery remains challenging and can be sensitive to annotation and modeling choices \citep{abumelha2025medical}.
As a result, higher atomic coverage may not always correspond to clinically sufficient information gathering.

% ICR depends on how the relevant evidence set E is defined and may inherit annotation bias.
ICR is defined with respect to a case specific relevant evidence set $E$.
In practice, multiple evidence subsets can justify the same diagnosis, and experts may disagree on what is necessary versus merely supportive.
Inter rater reliability in related clinical evaluation settings varies across attributes and rubrics, indicating that a single reference set can encode subjective decisions \citep{holderried2024language}.
Future work should report agreement statistics, test sensitivity to alternative definitions of $E$, and consider softer relevance modeling for borderline evidences.

%% file: latex/appendix/interactive_environment.tex
\subsection{Interactive Environment Details}

% This paragraph states the framework-level implementation choices and the stateless design motivation.
We implement an interactive diagnostic environment with three roles: a doctor agent, a simulated patient, and a \rev{simulated reporter}.
All roles are instantiated using the CAMEL multi-agent framework~\cite{li2023camel}.
Both the simulated patient and the \rev{simulated reporter} operate with a context window of 1 and a temperature of 0, resulting in stateless behavior with respect to dialogue history and deterministic decoding across all experiments.
This design choice reduces simulator drift over long conversations and improves stability across runs.

% This paragraph documents the patient prompt and the decoding configuration for reproducibility.
The patient prompt is shown below:

\begin{lstlisting}[
  basicstyle=\footnotesize\ttfamily,
  columns=flexible,
  breaklines=true,
  breakindent=0pt,
]
You are a patient undergoing a medical interview.
Your knowledge is strictly limited to the following list of indexed facts:
{patient_evidences}

Response protocols:
1. Analyze the doctor question and search your list for the specific item or items that contain the answer.

2. Format your output using two tags:
[REFERENCE] followed by the exact string or strings including the index from your list.
You may select up to two facts if necessary.
If no fact exists write N/A.
[RESPONSE] followed by a natural language answer derived strictly from the selected references.
Do not add outside information.

3. If the doctor question is not addressed by any fact in your list.
[REFERENCE] N/A
[RESPONSE] indicate that you are unsure or do not recall.
\end{lstlisting}

% This paragraph provides the reporter prompt used in our implementation.
The reporter prompt is shown below:

\begin{lstlisting}[
  basicstyle=\footnotesize\ttfamily,
  columns=flexible,
  breaklines=true,
  breakindent=0pt,
]
You are a specialized module named Measurement responsible for reporting test results to the physician.

You have access to the following list of indexed facts.
Physical examination and diagnostic test data
{examination_evidences}

Response protocols:
1. Search the provided list for all facts that are relevant to the doctor specific test request. Do not provide information that was not explicitly requested.

2. Return the relevant facts exactly as they appear in the source list including their indices.

3. If the requested test results are not found in the list assume the finding is non-significant and return Normal.
\end{lstlisting}

\subsection{Automatic Evaluation}

Following the original benchmarks, DiagnosisArena, RareArena, and ClinicalBench are \textbf{differential diagnosis} tasks where the model outputs five diseases ranked by likelihood.
AgentClinic-MedQA and Derm are \textbf{direct diagnosis} tasks that require a single diagnosis.

% This paragraph states the evaluation setup for direct diagnosis tasks and the judge output constraint.
For direct diagnosis tasks, we evaluate whether the doctor predicted diagnosis matches the reference diagnosis.
We use an LLM judge that outputs a binary decision with no additional text.
The judge prompt for direct diagnosis is shown below.

\begin{lstlisting}[
  basicstyle=\footnotesize\ttfamily,
  columns=flexible,
  breaklines=true,
  breakindent=0pt,
]
You determine whether the correct diagnosis and the doctor diagnosis refer to the same disease.
Respond only with Yes or No.

Correct diagnosis
{answer}

Doctor output
{diagnosis}

Are these the same disease? 
\end{lstlisting}

% This paragraph states the evaluation setup for differential diagnosis tasks and how scores relate to the main metric.
For differential diagnosis tasks, the doctor outputs five differential diagnoses.
We use an LLM judge to score each of the five items against the reference diagnosis using a three-level rubric.
A score of two indicates an exact match, a score of one indicates a broader category that contains the reference diagnosis, and a score of zero otherwise.
We compute success rate (SR) using top-1 accuracy by checking whether the first listed diagnosis receives a score of two.

% This paragraph provides the judge prompt for differential diagnosis tasks.
The judge prompt for differential diagnosis scoring is shown below:

\begin{lstlisting}[
  basicstyle=\footnotesize\ttfamily,
  columns=flexible,
  breaklines=true,
  breakindent=0pt,
]
You diagnose challenging cases.
You receive a student answer containing five differential diagnoses and a reference diagnosis.
Score each diagnosis using the rules below.

2 The student diagnosis exactly matches the reference diagnosis.
1 The student diagnosis is a broad category that includes the reference diagnosis.
0 The student diagnosis does not meet the criteria for 1 or 2.

Student answer
{student_answer}

Reference diagnosis
{final_diagnosis}

Output the scores in the format below and do not output anything else.
1 Disease 1 Name \boxed{score}
2 Disease 2 Name \boxed{score}
3 Disease 3 Name \boxed{score}
4 Disease 4 Name \boxed{score}
5 Disease 5 Name \boxed{score}
\end{lstlisting}

%% file: latex/appendix/evimed_construction_details.tex
\subsection{Data Sources and Sampling}

% This paragraph summarizes the composition of EviMed-1K and the high-level sampling policy.
EviMed-1K integrates five complementary sources that cover general medicine, specialty diagnosis, complex multi-specialty reasoning, rare diseases, and real-world clinical records.
We sample 200 cases from each source to form a 1,000-case evaluation set.

% This paragraph describes AgentClinic-MedQA and how it is used in EviMed.
\textbf{AgentClinic-MedQA}~\cite{schmidgall2024agentclinic} is a pure-text interactive clinical diagnosis dataset within the AgentClinic benchmark.
It contains 215 cases adapted from MedQA USMLE case challenges and rewritten into OSCE-style multi-round consultation scenarios.
The initial JSON cases were auto-filled using GPT-4 and then manually verified to ensure consistency and usability.
We randomly sample 200 cases to evaluate sequential information gathering and diagnosis under incomplete evidence.

% This paragraph describes the dermatology dataset composition and the inclusion policy.
\textbf{Derm}~\cite{johri2024craft} evaluates dermatology diagnosis with a public split (Derm-Public) and a clinician-authored split (Derm-Private).
Derm-Public contains 100 case-based questions collected from an online question bank.
Derm-Private contains 100 case-based questions newly written by three dermatologists with similar structure and different condition coverage to reduce leakage risk.
We include the full set of 200 cases to test detailed symptom inquiry in a specialized domain.

% This paragraph describes DiagnosisArena and the sampling choice for complex open-ended diagnosis.
\textbf{DiagnosisArena}~\cite{zhu2025diagnosisarena}is constructed from real-world case reports published in top-tier medical journals such as NEJM, The Lancet, and JAMA.
It extracts and structures diagnostic information including history, physical examination, and tests while removing treatment and prognosis content to reduce answer leakage.
The benchmark focuses on open-ended differential diagnosis without restricting candidates to a predefined list.
We randomly sample 200 cases from the dataset.

% This paragraph describes RareArena and the frequency-stratified sampling used to cover the long tail.
\textbf{RareArena}~\cite{zhao2025rarearena} is a large-scale rare disease benchmark built from PubMed Central (PMC) case reports and mapped to the Orphanet ORPHAcode system.
It includes Rare Disease Screening (RDS) and Rare Disease Confirmation (RDC) settings that reflect different stages of the diagnostic process.
To reflect the long-tail distribution and synonym variability, we sample 200 distinct diseases using a frequency-stratified scheme.
Specifically, diseases are drawn in a 2:2:1 ratio from low-frequency (appearing once), mid-frequency (appearing 2–5 times), and high-frequency (appearing more than 5 times) strata based on their occurrence counts in the corpus.

% This paragraph describes ClinicalBench and the coverage-oriented sampling across clinical departments.
\textbf{ClinicalBench}~\cite{yan2024clinicallab} is derived from de-identified electronic medical records with both structured and unstructured content.
It covers 24 clinical departments and primarily includes common diseases with clear diagnostic pathways that require multi-source clinical evidence.
The cases reflect realistic combinations of history, examination, imaging, and laboratory findings.
We sample 200 cases to ensure broad coverage across disease categories and specialties represented in the dataset.

\subsection{Automatic Evidence Construction}

% This paragraph provides the prompt used to automatically convert each case into atomic patient and examination evidence lists.
The prompt used to automatically construct atomic patient and examination evidences is shown below.

\begin{lstlisting}[
  basicstyle=\footnotesize\ttfamily,
  columns=flexible,
  breaklines=true,
  breakindent=0pt,
]
Break the following information into independent atomic facts.

Rules
- One piece of information per statement.
- Facts must be self-contained and non-overlapping.
- Do NOT add, infer, or normalize beyond the given text.
- Keep the original language of the input.
- Each fact string must start with an index such as "1. ", "2. ", and so on.
- Classify each fact into either patient_facts or exam_facts.
  - patient_facts include demographics, history, symptoms, complaints, and clinical presentation.
  - exam_facts include examinations, tests, laboratory results, and imaging studies.
- Do NOT duplicate facts across patient_facts and exam_facts.
  If a fact could belong to both, choose the best list and omit it from the other.
- If there is no content for a list, return an empty list.

Case information in JSON
{case_json}
\end{lstlisting}

%% file: latex/appendix/avg_turns.tex
\begin{table*}[t]
\caption{Average interaction turns (Turns), Information Coverage Rate (ICR, \%), and Turn Efficiency (Effi. = ICR / Turns) under different strategies across datasets.}
\label{tab:avg_turns}
\centering
\small
\begingroup
\setlength{\tabcolsep}{2.6pt}
\renewcommand{\arraystretch}{1.15}
\newcommand{\turncell}[1]{\makebox[2.8em][c]{#1}}
\newcommand{\icrcell}[1]{\makebox[2.8em][c]{#1}}
\newcommand{\ratecell}[1]{\makebox[2.8em][c]{#1}}
\newcommand{\efficell}[1]{\cellcolor{gray!15}\ratecell{#1}}
\resizebox{\textwidth}{!}{%
\begin{tabular}{l*{15}{c}}
\toprule
\textbf{Model}
& \multicolumn{3}{c}{\textbf{ClinicalBench}}
& \multicolumn{3}{c}{\textbf{Derm}}
& \multicolumn{3}{c}{\textbf{AgentClinic-MedQA}}
& \multicolumn{3}{c}{\textbf{DiagnosisArena}}
& \multicolumn{3}{c}{\textbf{RareArena}} \\
\cmidrule(lr){2-4}\cmidrule(lr){5-7}\cmidrule(lr){8-10}\cmidrule(lr){11-13}\cmidrule(lr){14-16}
& \turncell{\textbf{Turns}} & \textbf{ICR} & \cellcolor{gray!15}\ratecell{\textbf{Effi.}}
& \turncell{\textbf{Turns}} & \textbf{ICR} & \cellcolor{gray!15}\ratecell{\textbf{Effi.}}
& \turncell{\textbf{Turns}} & \textbf{ICR} & \cellcolor{gray!15}\ratecell{\textbf{Effi.}}
& \turncell{\textbf{Turns}} & \textbf{ICR} & \cellcolor{gray!15}\ratecell{\textbf{Effi.}}
& \turncell{\textbf{Turns}} & \textbf{ICR} & \cellcolor{gray!15}\ratecell{\textbf{Effi.}} \\
\midrule

\multicolumn{16}{c}{\textbf{Baseline}} \\
\midrule
GPT-5 & \turncell{12.1} & \icrcell{35.2} & \efficell{2.91} & \turncell{6.8} & \icrcell{54.9} & \efficell{8.07} & \turncell{7.4} & \icrcell{31.3} & \efficell{4.23} & \turncell{11.1} & \icrcell{55.4} & \efficell{4.99} & \turncell{12.3} & \icrcell{42.7} & \efficell{3.47} \\
GPT-5-mini & \turncell{12.5} & \icrcell{31.3} & \efficell{2.50} & \turncell{10.3} & \icrcell{51.7} & \efficell{5.02} & \turncell{10.9} & \icrcell{33.0} & \efficell{3.03} & \turncell{12.8} & \icrcell{47.2} & \efficell{3.69} & \turncell{12.9} & \icrcell{35.7} & \efficell{2.77} \\
DeepSeek-v3.2 & \turncell{12.7} & \icrcell{46.7} & \efficell{3.68} & \turncell{12.0} & \icrcell{77.0} & \efficell{6.42} & \turncell{11.3} & \icrcell{44.3} & \efficell{3.92} & \turncell{12.1} & \icrcell{63.4} & \efficell{5.24} & \turncell{12.6} & \icrcell{49.1} & \efficell{3.90} \\
GLM-4.6 & \turncell{8.6} & \icrcell{32.6} & \efficell{3.79} & \turncell{7.2} & \icrcell{65.7} & \efficell{9.13} & \turncell{7.5} & \icrcell{36.6} & \efficell{4.88} & \turncell{8.2} & \icrcell{52.1} & \efficell{6.35} & \turncell{9.2} & \icrcell{37.8} & \efficell{4.11} \\
Qwen2.5-72B & \turncell{8.2} & \icrcell{40.8} & \efficell{4.98} & \turncell{8.7} & \icrcell{67.5} & \efficell{7.76} & \turncell{7.9} & \icrcell{41.7} & \efficell{5.28} & \turncell{7.9} & \icrcell{55.5} & \efficell{7.03} & \turncell{8.0} & \icrcell{41.6} & \efficell{5.20} \\
Qwen2.5-32B & \turncell{7.9} & \icrcell{39.0} & \efficell{4.94} & \turncell{8.1} & \icrcell{71.6} & \efficell{8.84} & \turncell{7.1} & \icrcell{37.3} & \efficell{5.25} & \turncell{7.7} & \icrcell{57.0} & \efficell{7.40} & \turncell{8.3} & \icrcell{43.0} & \efficell{5.18} \\
Qwen2.5-7B & \turncell{6.3} & \icrcell{41.0} & \efficell{\textbf{6.51}} & \turncell{9.2} & \icrcell{71.9} & \efficell{7.82} & \turncell{7.8} & \icrcell{43.5} & \efficell{5.58} & \turncell{5.8} & \icrcell{53.1} & \efficell{9.16} & \turncell{6.5} & \icrcell{41.2} & \efficell{6.34} \\
Llama-3.1-8B & \turncell{8.7} & \icrcell{40.5} & \efficell{4.66} & \turncell{9.6} & \icrcell{74.4} & \efficell{7.75} & \turncell{8.7} & \icrcell{47.8} & \efficell{5.49} & \turncell{8.1} & \icrcell{57.3} & \efficell{7.07} & \turncell{8.9} & \icrcell{45.0} & \efficell{5.06} \\
Meditron3-8B & \turncell{16.0} & \icrcell{23.4} & \efficell{1.46} & \turncell{16.0} & \icrcell{39.7} & \efficell{2.48} & \turncell{16.0} & \icrcell{28.7} & \efficell{1.79} & \turncell{16.0} & \icrcell{33.7} & \efficell{2.11} & \turncell{16.0} & \icrcell{23.2} & \efficell{1.45} \\
Qwen2.5-3B & \turncell{8.3} & \icrcell{33.8} & \efficell{4.07} & \turncell{12.5} & \icrcell{74.1} & \efficell{5.93} & \turncell{9.7} & \icrcell{43.7} & \efficell{4.51} & \turncell{8.2} & \icrcell{47.7} & \efficell{5.82} & \turncell{9.4} & \icrcell{35.1} & \efficell{3.73} \\

\midrule
\multicolumn{16}{c}{\textbf{ReAct}} \\
\midrule
GPT-5-mini & \turncell{10.5} & \icrcell{33.5} & \efficell{3.19} & \turncell{8.1} & \icrcell{59.5} & \efficell{7.35} & \turncell{8.5} & \icrcell{33.2} & \efficell{3.91} & \turncell{10.6} & \icrcell{53.9} & \efficell{5.08} & \turncell{10.8} & \icrcell{39.3} & \efficell{3.64} \\
Qwen2.5-72B & \turncell{7.8} & \icrcell{38.7} & \efficell{4.96} & \turncell{8.1} & \icrcell{71.2} & \efficell{8.79} & \turncell{6.9} & \icrcell{39.5} & \efficell{5.72} & \turncell{7.1} & \icrcell{58.9} & \efficell{8.30} & \turncell{7.7} & \icrcell{47.4} & \efficell{6.16} \\
Qwen2.5-7B & \turncell{4.8} & \icrcell{29.8} & \efficell{6.21} & \turncell{4.6} & \icrcell{63.5} & \efficell{\textbf{13.80}} & \turncell{4.3} & \icrcell{35.8} & \efficell{\textbf{8.33}} & \turncell{4.6} & \icrcell{45.1} & \efficell{\textbf{9.80}} & \turncell{4.8} & \icrcell{29.8} & \efficell{6.21} \\

\midrule
\multicolumn{16}{c}{\textbf{SC}} \\
\midrule
GPT-5-mini & \turncell{13.5} & \icrcell{41.5} & \efficell{3.07} & \turncell{10.8} & \icrcell{70.9} & \efficell{6.56} & \turncell{10.7} & \icrcell{40.2} & \efficell{3.76} & \turncell{13.2} & \icrcell{60.8} & \efficell{4.61} & \turncell{13.5} & \icrcell{46.2} & \efficell{3.42} \\
Qwen2.5-72B & \turncell{9.4} & \icrcell{43.2} & \efficell{4.60} & \turncell{10.4} & \icrcell{77.9} & \efficell{7.49} & \turncell{8.4} & \icrcell{43.6} & \efficell{5.19} & \turncell{8.6} & \icrcell{63.4} & \efficell{7.37} & \turncell{9.1} & \icrcell{50.1} & \efficell{5.51} \\
Qwen2.5-7B & \turncell{5.9} & \icrcell{32.5} & \efficell{5.51} & \turncell{5.8} & \icrcell{62.8} & \efficell{10.83} & \turncell{4.9} & \icrcell{36.0} & \efficell{7.35} & \turncell{5.7} & \icrcell{53.6} & \efficell{9.40} & \turncell{5.7} & \icrcell{37.0} & \efficell{\textbf{6.49}} \\

\midrule
\multicolumn{16}{c}{\textbf{REFINE}} \\
\midrule
GPT-5-mini & \turncell{15.7} & \icrcell{43.5} & \efficell{2.77} & \turncell{14.7} & \icrcell{76.7} & \efficell{5.22} & \turncell{13.5} & \icrcell{45.5} & \efficell{3.37} & \turncell{15.3} & \icrcell{64.6} & \efficell{4.22} & \turncell{15.5} & \icrcell{51.8} & \efficell{3.34} \\
Qwen2.5-72B & \turncell{13.9} & \icrcell{\textbf{51.1}} & \efficell{3.68} & \turncell{14.5} & \icrcell{\textbf{80.8}} & \efficell{5.57} & \turncell{12.6} & \icrcell{\textbf{53.9}} & \efficell{4.28} & \turncell{13.2} & \icrcell{\textbf{73.8}} & \efficell{5.59} & \turncell{14.1} & \icrcell{\textbf{59.9}} & \efficell{4.25} \\
Qwen2.5-7B & \turncell{6.1} & \icrcell{35.3} & \efficell{5.79} & \turncell{8.1} & \icrcell{67.1} & \efficell{8.28} & \turncell{6.6} & \icrcell{38.6} & \efficell{5.85} & \turncell{5.8} & \icrcell{50.8} & \efficell{8.76} & \turncell{6.0} & \icrcell{38.2} & \efficell{6.37} \\

\bottomrule
\end{tabular}%
}
\endgroup
\end{table*}

Table~\ref{tab:avg_turns} reports the average number of interaction turns.
We additionally report turn efficiency as Effi. = ICR / Turns to quantify evidence acquisition per interaction step.

Across datasets, some models with stronger general reasoning ability, such as GPT-5 and GLM-4.6, exhibit longer dialogues, yet the resulting ICR gains remain limited.
In contrast, the Qwen series often achieves relatively high ICR with fewer turns.
This yields consistently higher efficiency, suggesting that these models ask more targeted questions and extract salient evidence earlier.
Meditron3-8B frequently reaches the maximum turn budget but attains low ICR, indicating limited capability in interactive information collection.

Among strategies, ReAct typically improves efficiency even when the absolute ICR gain is modest.
REFINE more often increases ICR by extending the interaction, but this additional turn cost can lead to lower efficiency than ReAct.

%% file: latex/appendix/avg_turn_time.tex
\begin{table}[!t]
  \centering
  \caption{Average per-turn latency relative to Baseline.}
  \label{tab:turn_time_multiplier}
  \begin{tabular}{l r}
    \hline
    Strategy & Avg.\ Turn Latency \\
    \hline
    Baseline & $\times 1.00$ \\
    ReAct & $\times 3.85$ \\
    SC & $\times 10.22$ \\
    REFINE & $\times 16.48$ \\
    \hline
  \end{tabular}
\end{table}

We report per-turn latency for the doctor agent as a proxy for interactive efficiency.
A turn is timed from when the doctor receives the turn-level visible context to when the doctor finishes generating the next action.
This proxy reflects user-perceived responsiveness in deployment settings.

Experiments are run locally on a single NVIDIA A6000 GPU using vLLM with Qwen2.5-7B.
All numbers are averaged over five datasets.

% This paragraph interprets the results and states the implication.
Table~\ref{tab:turn_time_multiplier} shows that strategies with explicit reasoning traces or additional internal steps increase per-turn latency.
These results highlight a practical trade-off between accuracy improvements and runtime cost.

%% file: latex/appendix/prompts.tex
This appendix lists all prompts used by different strategies.

\paragraph{Baseline.}
The Baseline strategy includes only a doctor agent.

\begin{lstlisting}[
  basicstyle=\footnotesize\ttfamily\color{blue!65!black},
  columns=flexible,
  breaklines=true,
  breakindent=0pt,
]
# Role: Doctor
You are a licensed physician conducting a medical consultation.
{task_description}

Your objective is to efficiently gather information and request necessary clinical examinations or laboratory tests to enable a subsequent diagnostic analysis.
You have access to a Medical Analyst who can retrieve specific test results upon request.

You must adhere to the following operational constraints:
1. Efficiency: Gather sufficient information in as few turns as possible.
2. Turn Limit: You strictly cannot exceed {max_turns} total turns.
3. No Repetition: Never ask a question or request a test that has already been covered.
4. Atomic Inquiries: Each question must address a single, specific topic.
   For example, ask "What are your symptoms?" and never combine multiple questions.

In every turn, you execute one of the following actions in the corresponding format:
1. [QUERY] followed by your atomic question to the patient.
2. [TEST] followed by one specific examination or diagnostic test request to the Medical Analyst.
3. [DIAGNOSIS] followed by {task_output_format}

Once you have gathered sufficient evidence, ensure your diagnosis is final.

\end{lstlisting}

\paragraph{ReAct.}
The ReAct strategy also includes only a doctor agent.

\begin{lstlisting}[
  basicstyle=\footnotesize\ttfamily\color{green!60!black},
  columns=flexible,
  breaklines=true,
  breakindent=0pt,
]
# Role: Doctor (ReAct)
You are a licensed physician conducting a medical consultation.
{task_description}

Your objective is to efficiently gather information and request necessary clinical examinations or laboratory tests to enable a subsequent diagnostic analysis.
You have access to a Medical Analyst who can retrieve specific test results upon request.

You must adhere to the following operational constraints:
1. Efficiency: Gather sufficient information in as few turns as possible.
2. Turn Limit: You strictly cannot exceed {max_turns} total turns.
3. No Repetition: Never ask a question or request a test that has already been covered.
4. Atomic Inquiries: Each question must address a single, specific topic.
   For example, ask "What are your symptoms?" and never combine multiple questions.

In every turn, you must follow a strict Reasoning-then-Acting process using the following format exactly.

[THOUGHT] <Your Clinical Reasoning>
  - Analyze the current clinical picture, identify critical information gaps, and articulate step-by-step reasoning to justify your next action.

Execute exactly ONE of the following three commands based on your thought process.
  - [QUERY] followed by your atomic question to the patient.
  - [TEST] followed by one specific examination or diagnostic test request to the Measurement module.
  - [DIAGNOSIS] followed by {task_output_format}

Once you have gathered sufficient evidence, ensure your diagnosis is final.

\end{lstlisting}

\paragraph{SC.}
The SC strategy includes an Information Collector, an Evidence Organizer, and a Diagnosis Reasoner.

\begin{lstlisting}[
  basicstyle=\footnotesize\ttfamily\color{teal!65!black},
  columns=flexible,
  breaklines=true,
  breakindent=0pt,
]
# Role: Information Collector
You are a licensed physician conducting a medical consultation.
{task_description}

Your objective is to efficiently gather information and request necessary clinical examinations or laboratory tests to enable a subsequent diagnostic analysis.
You have access to a Medical Analyst who can retrieve specific test results upon request.

You must adhere to the following operational constraints:
1. Efficiency: Gather sufficient information in as few turns as possible.
2. Turn Limit: You strictly cannot exceed {max_turns} total turns.
3. No Repetition: Never ask a question or request a test that has already been covered.
4. Atomic Inquiries: Each question must address a single, specific topic.
   For example, ask "What are your symptoms?" and never combine multiple questions.

In every turn, you must follow a strict Reasoning-then-Acting process using the following format exactly.

[THOUGHT] <Your Clinical Reasoning>
  - Analyze the current clinical picture, identify critical information gaps, and articulate step-by-step reasoning to justify your next action.

Execute exactly ONE of the following three commands based on your thought process.
  - [QUERY] followed by your atomic question to the patient.
  - [TEST] followed by one specific examination or diagnostic test request to the Measurement module.
  - [FINISH] use this command ONLY when you believe you have gathered all necessary information to form a conclusive diagnosis.
    You do not need to provide a diagnosis.

Once you issue the [FINISH] command, the consultation ends immediately.

\end{lstlisting}

\begin{lstlisting}[
  basicstyle=\footnotesize\ttfamily\color{brown!70!black},
  columns=flexible,
  breaklines=true,
  breakindent=0pt,
]
# Role: Evidence Organizer
You are a professional medical documentarian and clinical scribe.
Your objective is to synthesize the dialogue between a doctor and a patient into a high-fidelity structured medical summary.

Core Principles:
1. Strict Adherence: You must NOT invent, infer, or hallucinate any information not explicitly present in the dialogue.
2. Precision: Retain all precise measurements, dates, dosages, and technical medical terms exactly as stated.
3. Objectivity: Maintain a professional, clinical tone throughout the summary.

Output Process:
You must follow this process and use the following format exactly:

[THOUGHT]
Analyze the dialogue to extract key clinical facts and reasoning. Plan how to organize these details logically, ensuring no critical information is overlooked.

[SUMMARY]
Generate a professional, structured clinical note. You should organize the content in the format that best fits the case context, ensuring the summary is comprehensive, coherent, and clinically accurate.

\end{lstlisting}

\begin{lstlisting}[
  basicstyle=\footnotesize\ttfamily\color{violet!70!black},
  columns=flexible,
  breaklines=true,
  breakindent=0pt,
]
# Role: Diagnosis Reasoner
You are a senior diagnostic physician specializing in complex differential diagnosis.
{task_description}
Your objective is to analyze the provided structured clinical summary to formulate a precise diagnosis.
You must follow a strict reasoning process and use the following format exactly:

[THOUGHT] <Your Clinical Reasoning>
   -Perform a comprehensive clinical analysis of the summary.

[DIAGNOSIS]
   - Provide the {task_output_format}.
\end{lstlisting}

\paragraph{REFINE.}
The REFINE strategy includes an Information Collector, an Evidence Organizer, a Diagnosis Reasoner, and a Diagnosis Verifier.
The prompts for the Information Collector, Evidence Organizer, and Diagnosis Reasoner are identical to those used in SC.

\begin{lstlisting}[
  basicstyle=\footnotesize\ttfamily\color{black!75},
  columns=flexible,
  breaklines=true,
  breakindent=0pt,
]
# Role: Diagnosis Verifier
You are a Clinical Diagnostic Supervisor.
{task_description}
Your objective is to evaluate sufficiency of the diagnosis provided by the physician, based strictly on the available case summarized information.

Evaluation Criteria:
1. Data Sufficiency: Determine if the current information is actually sufficient to form a conclusive diagnosis.
2. Turn Limit Override: If the maximum turn limit has been reached (like "Turn 12/12"). You must force a decision (PASS or REJECT) based on the best possible interpretation of existing data.

You must follow this process and use the following format exactly:

[THOUGHT] <Your Analysis>
   - Identify if any "Red Flag" symptoms or critical tests are missing that prevent a safe diagnosis.

[DECISION] <Status>
   - Output "PASS" if the diagnosis is sufficient.
   - Output "INCOMPLETE" if the diagnosis is premature because critical clinical information is missing. (Requires the Physician to return to the patient to gather more data. Only valid if not reach maximum turn).

[FEEDBACK] <Guidance>
   - If PASS: Leave this section empty.
   - If INCOMPLETE: Specify exactly what critical information (e.g., specific missing lab test, biopsy, or history) is required to form a valid diagnosis.
\end{lstlisting}